\documentclass[10pt,leqno]{amsart}

\usepackage[utf8]{inputenc}
\usepackage[T1]{fontenc}
\usepackage{amsmath,amssymb,amsthm,amsfonts}
\usepackage{graphicx}
\usepackage{xcolor}
\usepackage{booktabs}
\usepackage{array}
\usepackage{makecell}
\usepackage{booktabs}
\usepackage{multirow}
\usepackage{rotating}
\usepackage{ragged2e} 
\usepackage{makecell}
\usepackage{float}
\usepackage{textcomp}
\usepackage{url}
\usepackage{verbatim}
\usepackage{multicol}
\usepackage{microtype}
\usepackage{tikz}
\usetikzlibrary{patterns}
\usepackage{pgfplots}
\usepackage{pgfplotstable}
\pgfplotsset{compat=1.7}
\usepackage{subfig} 
\usepackage{algorithm}
\usepackage{algorithmic}
\usepackage{csquotes}
\usepackage[numbers,sort&compress]{natbib}          
\usepackage{epstopdf}       
\usepackage[strings]{underscore} 
\usepackage[hidelinks]{hyperref} 

\usepackage[utf8]{inputenc}
\numberwithin{equation}{section}
\DeclareUnicodeCharacter{0080}{ }
\UseRawInputEncoding
\usepackage[utf8]{inputenc}
\usepackage[T1]{fontenc}
\usepackage{microtype}  
\usepackage[english]{babel}

\begin{document}
\title{ Colorectal Cancer Histopathological Grading using Multi-Scale Federated Learning} 

\author[Arafath]{Md. Ahasanul Arafath}
\address{Department of Computer Science and Engineering, European University of Bangladesh}
\email{ahasanul.arafath@gmail.com}

\author[]{⁠Abhijit Kumar Ghosh}
\address{Department of Computer Science and Engineering , BRAC University,Dhaka, Bangladesh}
\email{abhijit.kumar.ghosh.77880@gmail.com}

\author[]{Md Rony Ahmed}
\address{Department of Computer Science and Engineering , Daffodil International University, Dhaka, Bangladesh}
\email{abhijit.kumar.ghosh.77880@gmail.com}

\author[Afroz]{Sabrin Afroz}
\address{Department of Institute of Information Technology, University of Dhaka, Bangladesh}
\email{sabrinafrozsathi@gmail.com}

\author[]{Minhazul Hosen}
\address{Department of Computer Science and Engineering , Daffodil International University, Dhaka, Bangladesh}
\email{njniloy113240@gmail.com}

\author[]{Md. Hasan Moon}
\address{Department of Computer Science and Engineering , Daffodil International University, Dhaka, Bangladesh}
\email{hasanmoon9768@gmail.com}

\author[]{⁠Md Tanzim Reza}
\address{Department of Computer Science and Engineering , BRAC University,Dhaka}
\email{tanzim.reza@bracu.ac.bd}

\author[Alam]{Md.~Ashad Alam}
\address{Ochsner Center for Outcomes Research, Ochsner Research, New Orleans, LA 70121, USA}
\email{mdashad.alam@ochsner.org}

\begin{abstract}
Colorectal cancer (CRC) grading is a critical prognostic factor but remains hampered by inter-observer variability and the privacy constraints of multi-institutional data sharing. While deep learning offers a path to automation, centralized training models conflict with data governance regulations and neglect the diagnostic importance of multi-scale analysis. In this work, we propose a scalable, privacy-preserving federated learning (FL) framework for CRC histopathological grading that integrates multi-scale feature learning within a distributed training paradigm. Our approach employs a dual-stream ResNetRS50 backbone to concurrently capture fine-grained nuclear detail (at 320×320 pixels) and broader tissue-level context (at 224×224 pixels). This architecture is integrated into a robust FL system stabilized using FedProx to mitigate client drift across heterogeneous data distributions from multiple hospitals. Extensive evaluation on the CRC-HGD dataset demonstrates that our framework achieves an overall accuracy of 83.5\%, outperforming a comparable centralized model (81.6\%). Crucially, the system excels in identifying the most aggressive Grade III tumors with a high recall of 87.5\%, a key clinical priority to prevent dangerous false negatives. Performance further improves with higher magnification, reaching 88.0\% accuracy at 40×. These results validate that our federated multi-scale approach not only preserves patient privacy but also enhances model performance and generalization. The proposed modular pipeline, with built-in preprocessing, checkpointing, and error handling, establishes a foundational step towards deployable, privacy-aware clinical AI for digital pathology.. 
\end{abstract} 

\maketitle

\section{Introduction}
\label{sec:related}
Colorectal cancer (CRC) accounts for nearly 2 million new
cases annually and is the second leading cause of cancer-
related deaths globally. Prognosis and therapeutic planning
are heavily influenced by tumor grade, reflecting the degree
of cellular differentiation. Accurate grading allows clinicians
to stratify patients into appropriate treatment pathways,
ranging from surgical resection to aggressive
chemoradiotherapy \cite{afroz2024multi, Sung2021}. In a most important sense, robust
grading is the key to uncovering those tumors, which are badly
differentiated (Grade III), that require the most intensive
treatment and hence rapid reaction.
Despite its clinical significance, manual histopathological
grading is subject to variability due to subjective
interpretation, workload pressures, and a lack of
standardization across institutions, with inter-observer
agreement often challenging the distinction between intermediate and high-grade tumors. With the rapid growth of
digital pathology, machine learning has emerged as a
powerful tool to automate cancer grading and reduce
diagnostic burden \cite{Janowczyk2016, Richfield-17,Zhu-24}. However, most prior works assume
access to centralized, large-scale datasets, which conflicts
with stringent privacy regulations such as HIPAA (U.S.) and
GDPR (EU). Moreover, pathology data utilized in current
federated learning research just neglect the fact that the multi-
scale context is very important, leading to models that may not
include the necessary diagnostic features.
Federated learning (FL) offers a paradigm shift, allowing
multiple institutions to collaboratively train AI models
without exchanging raw data \cite{Yang2019,Hassan-21}. In parallel, multi-scale imaging is central to pathology: high magnifications (40$\times$) capture nuclear atypia and mitotic figures, while lower magnifications (10$\times$–20$\times$) provide tissue- and gland-level context \cite{Madabhushi2016}. Integrating these complementary scales within a federated framework opens the door to more robust and generalizable grading models \cite{Auliah2021,Ashad-10, Ashad-11, Ashad-13}.

In this work, we propose a federated multi-scale learning framework for CRC histopathological grading that addresses both privacy and diagnostic accuracy. Our main contributions are:
\begin{itemize}
    \item {Dual-stream backbone:} A dual-stream ResNetRS50 design that captures fine-grained nuclear detail alongside broader morphological context.
    \item {Federated stability:} An FL system augmented with FedProx to improve stability and convergence across heterogeneous institutional clients.
    \item {Comprehensive evaluation:} Experiments on the CRC-HGD dataset demonstrating strong Grade~III recall and magnification-dependent gains.
    \item {Scalable pipeline:} A modular, reproducible pipeline with automatic checkpointing, error handling, and flexible dataset integration to facilitate deployment.
\end{itemize}

\section{Related work}
\label{Sec:rew}
\subsection{Deep Learning in Histopathology}
Deep learning, particularly convolutional neural networks (CNNs), has transformed digital pathology, delivering state-of-the-art performance in tumor detection, survival prediction, and grading \cite{Coudray2018}. Foundational work by Janowczyk and Madabhushi \cite{Janowczyk2016} established the feasibility of automated histopathological analysis and helped catalyze computational pathology. Multimodal fusion has been introduced integrates histopathology with genomics for improved clinical insight. Despite these advances, most systems rely on centralized training, raising privacy and governance concerns.

\subsection{Federated Learning in Medical Imaging}
FL has emerged as a promising paradigm for privacy-preserving model development in healthcare. Early feasibility studies in multi-institutional settings showed that robust models can be trained without sharing raw data \cite{Sheller2018}. Recent educational applications of FL, such as those by Farooq et al. (2024), demonstrate its predictive strength across diverse classifiers while maintaining privacy, reinforcing its cross-domain viability for sensitive data environments \cite{Farooq2024}. Similarly, Fachola et al. (2023) implemented a cross-silo federated neural network for dropout prediction across educational institutions, demonstrating that FL can match centralized performance while preserving institutional privacy and minimizing data movement \cite{Fachola2023}. Xu \textit{et al.} surveyed FL for healthcare informatics, highlighting challenges such as heterogeneity and communication efficiency \cite{Xu2020}. However, applications to digital pathology remain comparatively limited particularly for cancer grading tasks that demand explicit multi-scale feature integration.  Recent work by Latif \& Zhai (2025) demonstrates the effectiveness of federated learning for privacy-preserved automated scoring in educational research, using LoRA-based model adaptation and adaptive aggregation strategies \cite{latif2025privacypreservedautomatedscoringusing}. Their findings, though focused on educational data, offer architectural parallels and privacy guarantees that are highly relevant to medical imaging tasks, particularly in digital pathology where data heterogeneity and institutional silos are prevalent.


\subsection{Multi-Scale Learning in Pathology}
Pathologists routinely examine tissue at multiple magnifications, moving between low-power and high-power fields to capture both contextual and cellular cues. Inspired by this practice, multi-scale CNNs have demonstrated notable gains in detection and classification \cite{Wang2020}. For instance, Hou \textit{et al.} \cite{lehou2016patch} leveraged hierarchical context with patch-based models for whole-slide classification. Our approach extends this line of work by coupling multi-scale representation learning with federated collaboration, a combination that remains underexplored yet highly relevant for privacy-preserving digital pathology.

\section{Materials and Methods}\label{sec:methods}

\begin{figure*}[tbp]
 \centering
\includegraphics[width=13cm]{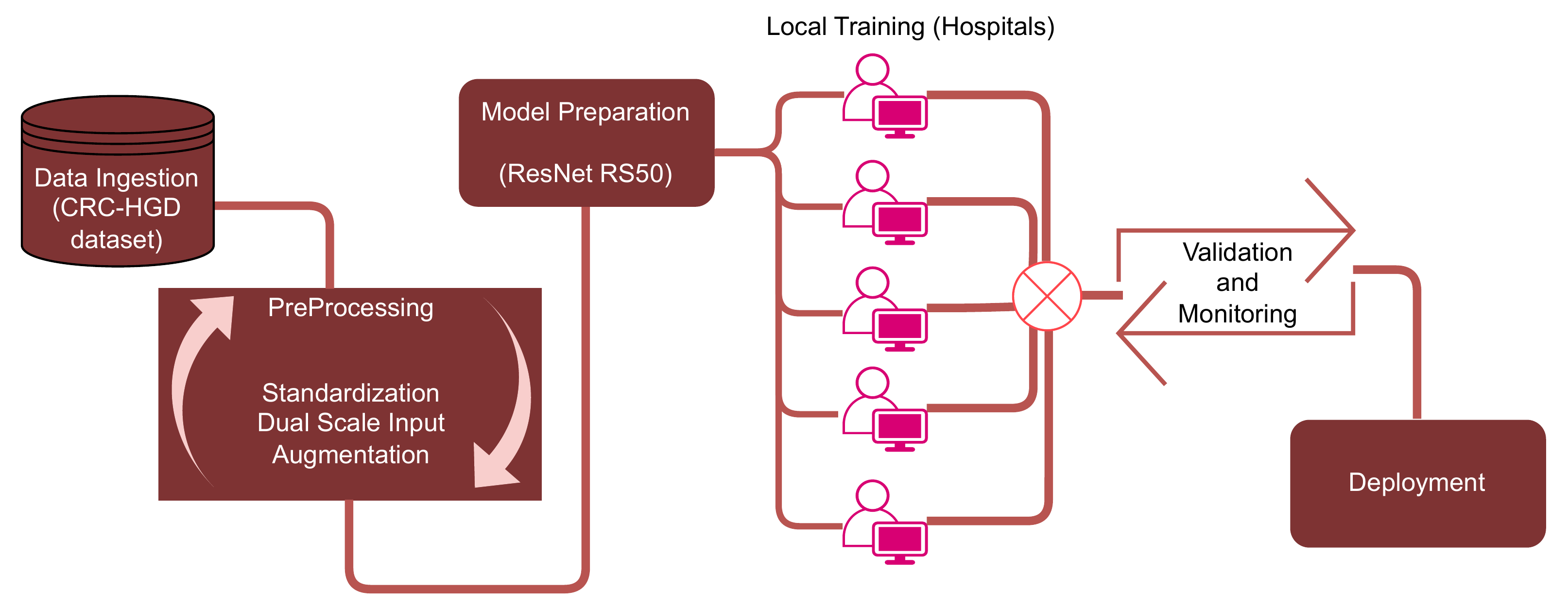}
 \caption{System architecture for multi-scale colorectal cancer grading using dual-stream processing.}
 \label{fig:workflow_multi_scale_federate_learning}
 \end{figure*}

\subsection{Dataset: CRC-HGD}
This study uses the CRC-HGD v2 dataset \cite{Amjadi2024, Alam-18a} that consist of 1{,}899 histopathological images of colorectal cancer annotated into three histological grades and provided at four magnifications (4$\times$, 10$\times$, 20$\times$, 40$\times$), mirroring routine clinical review.

\begin{figure}[H]
\centering
\includegraphics[width=.9\linewidth]{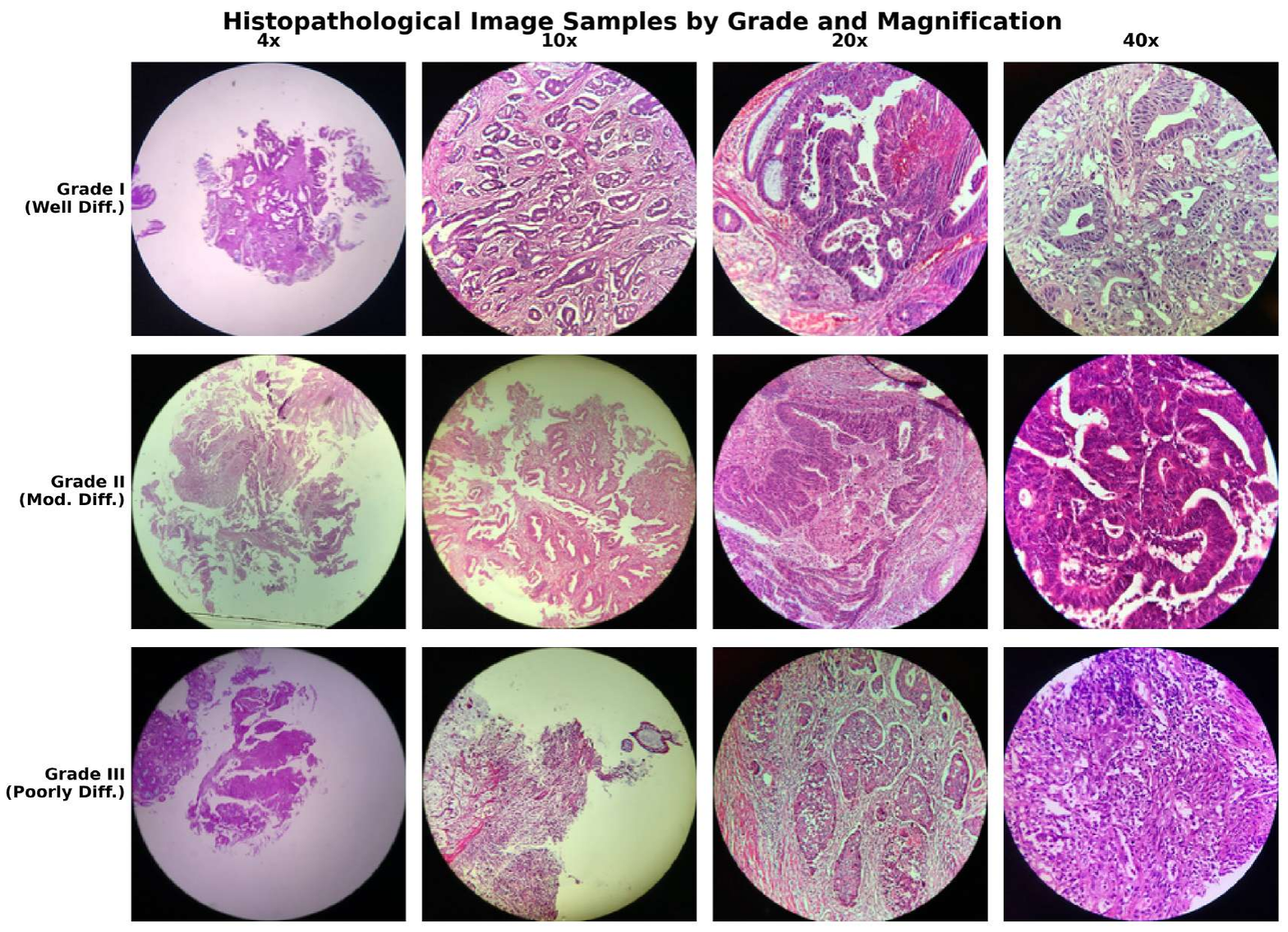}
\caption{Representative histopathology samples showing morphological variations across tumor grades and magnifications}
\label{fig:sample_gallery}
\end{figure}

\begin{itemize}
    \item {Grade I (Well differentiated):} Tumor cells closely 
resemble normal glandular structures (Lowest aggressiveness).

    \item {Grade II (Moderately differentiated):}  Represents the 
majority of cases, showing intermediate architectural 
disorganization.

    \item {Grade III (Poorly differentiated):} Cells show the least 
resemblance to normal tissue, often with solid growth patterns 
and minimal glandular formation (Highest aggressiveness). 
\end{itemize}
\subsection{Preprocessing} 
A comprehensive preprocessing pipeline was designed to ensure robust federated training and compatibility with the dual-scale network architecture \cite{Alam-16a, Alam-16b}. The main steps are as follows:

\begin{enumerate}
    \item {Image Standardization:} Raw histopathological images were normalized using stain-standardization methods such as Macenko~\cite{Macenko2009} or Reinhard~\cite{Reinhard2001}. This mitigates variations in hematoxylin and eosin (H\&E) staining and scanner hardware across institutions, thereby improving model robustness.

    \item {Patch Extraction:} 
    Whole-slide images (WSIs) were tiled into fixed-size patches. Non-informative background regions were excluded using Otsu’s thresholding on tissue masks~\cite{Otsu1979}, ensuring that only diagnostically relevant tissue contributed to training.

    \item {Dual-Scale Input Generation:} For each patch, two resolutions were produced to support the dual-stream backbone:
    \begin{itemize}
        \item Small scale (224 $\times$ 224 pixels): captures morphological context.
        \item Large scale (320 $\times$ 320 pixels): highlights cellular and nuclear detail.
    \end{itemize}

    \item {Artifact Removal:} Low-quality patches with blur, pen marks, or tissue folds were detected using Laplacian-based focus metrics and morphological heuristics \cite{Tellez2019} and removed to maintain dataset integrity. 

    \item {Duplicate Removal:} Redundant patches from the same client were detected using perceptual hashing and eliminated to reduce over representation.

    \item {Color Augmentation:} Stain invariance was improved through color jittering (brightness, contrast, hue, and saturation adjustments), simulating inter-lab variability.

    \item {Class-Balanced Sampling:} Since Grade III tumors were underrepresented, class-balanced sampling ensured adequate inclusion of minority classes during training, preventing model bias.

    \item {MixUp Augmentation:} MixUp~\cite{Zhang2017} was applied to improve generalization. Given two images $x_i, x_j$ with labels $y_i, y_j$, synthetic samples were generated as:
    \begin{equation}
        \tilde{x} = \lambda x_i + (1-\lambda)x_j,
    \end{equation}
    \begin{equation}
        \tilde{y} = \lambda y_i + (1-\lambda)y_j,
    \end{equation}
    where $\lambda \sim \text{Beta}(\alpha,\alpha)$ with $\alpha=0.2$. This strategy enforces smoother decision boundaries and mitigates class imbalance.
\end{enumerate}
\subsection{Network Architecture}
The framework uses a dual-stream backbone with two ImageNet-pretrained ResNetRS50 encoders \cite{Bello2021} operating at different scales to capture both tissue context and cellular detail essential for colorectal cancer grading.

\begin{enumerate}
    \item {Dual-Stream ResNetRS50 Encoders:} Each stream uses a ResNetRS50 pre-trained on ImageNet (via the timm library).
    \begin{itemize}
        \item {Coarse stream (320 $\times$ 320 pixels) :} captures glandular and stromal patterns. 
        \item {Fine stream (224 $\times$ 224 pixels):} highlights nuclear morphology and mitotic activity.
    \end{itemize}
    \item {Feature Fusion and Classification Head :}
    The two feature vectors are concatenated to form a joint representation:
    \begin{equation}
    f = [\,f_c \,\|\, f_f\,],
    \end{equation}
    which is passed through a fully connected classification head comprising dense layers with ReLU activation and dropout regularization. The final output layer employs Softmax to predict class probabilities for grades I – III
\end{enumerate}


\subsection{Federated Setup}

A federation of four hospital clients was simulated to mimic real-world collaboration. Training ran for 10 rounds, with each client performing three local epochs, that balance local refinement and communication efficiency. Adam optimizer (lr $= 3\times10^{-4}$, weight decay $= 1 \times 10^{-4}$) and the standard FedAvg algorithm were used for global aggregation.
\begin{equation}
w_{t+1} = \sum_{k=1}^K \frac{n_k}{n} \, w_k, 
\quad n=\sum_{k=1}^K n_k,
\end{equation}
where $w_k$ denotes the local weights and $n_k$ the local sample size.  

To address client drift from non-IID distributions, FedProx~\cite{Macenko2009} was integrated by adding a proximal term to each local objective:
\begin{equation}
\min_{w} \; F_k(w) + \frac{\mu}{2}\lVert w-w_t \rVert^2, \qquad \mu=0.01.
\end{equation}

Zhang et al. (2024) introduced a clustered FL framework that groups clients by model similarity, maintaining FedAvg-level efficiency while improving personalization and stability in non-IID settings \cite{Zhang2024}.
\subsection{Pipeline Design}
The proposed system follows a modular pipeline architecture composed of several core components: data ingestion, preprocessing, local training, aggregation, validation, and monitoring. This modularity ensures flexibility and scalability, thereby facilitating deployment in both research and clinical settings \cite{Alam-18C, Alam-21}.

To support diverse data-sharing practices across institutions, the pipeline accepts two input formats: structured folders and RAR archives. This dual-mode ingestion simplifies interoperability between sites with different data storage conventions. Centralized configuration management governs all modules, ensuring standardization of experimental settings, reproducibility, and streamlined deployment \cite{Ashad-14, Ashad-15, alam2014kernel}.

For reliability, the pipeline integrates automatic checkpointing and complete metadata snapshots, enabling efficient traceability of models and experiments. Robust error-handling mechanisms are embedded at every stage, ensuring uninterrupted operation during federated \cite{Ashad-08, Alam-19}. Together, these design features provide a high-performance, fault-tolerant, and easily distributable infrastructure tailored to the unique requirements of federated digital pathology.

\section{Results and Analysis}
\label{Sec:res}
\subsection{Overall Performance}
As a result, the proposed federated learning system achieved a test accuracy of {83.5\%}, with the best validation accuracy of {77.6\%} observed at round 9. Such outcomes indicate that the model demonstrated stable convergence throughout the federated rounds, reflecting the robustness of the FedProx-based FedAvg aggregation strategy in handling client drift, which is often caused by heterogeneous hospital data distributions. To ensure methodological rigor and comparability, our evaluation strategy aligns with recent federated learning benchmarking standards outlined by Chai et al. (2024), which emphasize realistic distributed workloads, fairness, and privacy-aware performance metrics \cite{Chai2024}. Furthermore, no signs of system overfitting were observed, underscoring the reliability of the modular pipeline design, which incorporated features such as automated checkpointing, metadata snapshots, and comprehensive error handling.  

The close agreement between validation and test performance further highlights the model’s generalization ability beyond the training distribution. This characteristic is particularly crucial for real-world deployment across multiple hospital scenarios in clinical practice, where the ability to maintain consistency under varying data sources is essential.
    \begin{table*}

    \caption{Overall performance metrics of the federated learning model on colorectal cancer grading}
    \begin{center}
    \resizebox{\columnwidth}{!}{
    \begin{tabular}{ccc}
    \hline
    \textbf{Metric} & \textbf{Value} & \textbf{Significance} \\
    \hline
    Test Accuracy & 83.5\% & Comparable to human pathologists \\
    \hline
    Best Validation Accuracy & 77.6\% (Round 9) & Good convergence pattern \\
    \hline
    Generalization & Test \textgreater Validation & minimum overfitting issue \\
    \hline
    Macro F1-Score & 83.3\% & Balanced across classes \\
    \hline
    Weighted F1-Score & 83.7\% & Accounts for class imbalance \\
    \hline
    \end{tabular}
    }
    \label{tab1}
    \end{center}
     \end{table*}

    \begin{figure}
    \centering
    \includegraphics[width=0.65\linewidth]{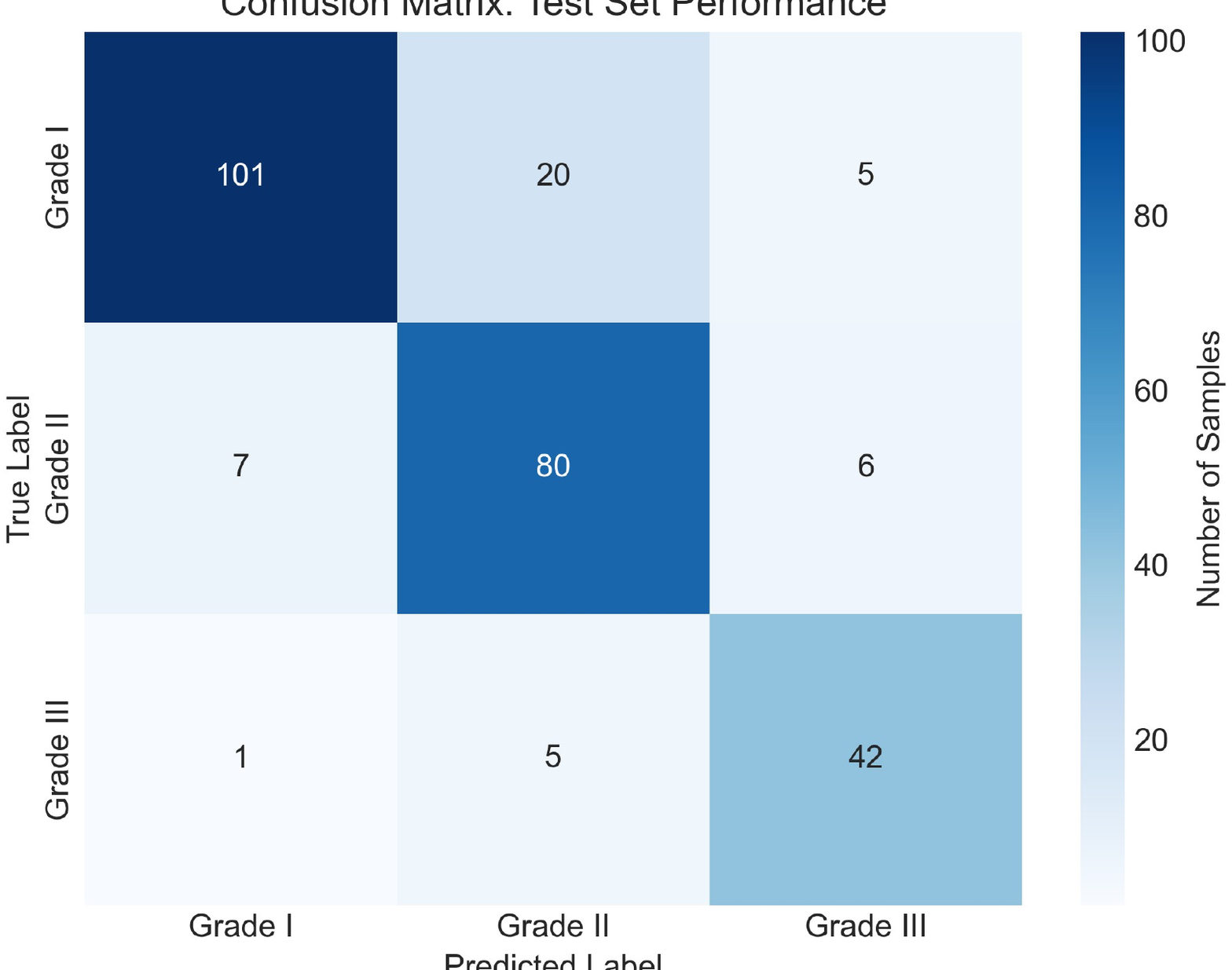}
    \caption{Confusion matrix showing the classification performance of the model on the test set for tumor Grades I, II, and III. The diagonal elements (highlighted) represent the correct predictions}
    \label{fig:confusion_matrix}
    \end{figure}

\subsection{Federated vs. Centralized Learning: A Comparative Analysis}
To validate the credibility of our federated learning approach against traditional centralized training, we conducted a comparative analysis using identical model architectures and training configurations. Remarkably, the FL framework achieved superior overall performance (83.5\% vs 81.6\%) despite the privacy-preserving distributed training paradigm. The FL model demonstrated particular strength at high magnifications (88.0\% at 40x) crucial for nuclear feature analysis, while maintaining balanced performance across all tumor grades. This performance advantage, coupled with FL's inherent privacy benefits, establishes federated learning as both a practical and competitive approach for multi-institutional clinical AI collaboration, effectively addressing the critical trade-off between data privacy and model performance.
\begin{table}[H]
\centering
\caption{Performance comparison between federated learning and centralized training approaches}
\begin{tabular}{cccc}
\hline
\textbf{Metric} & \textbf{Federated Learning} & \textbf{Non-FL} & \textbf{Difference} \\ 
\hline
Overall Accuracy      & 83.5\% & 81.6\% & +1.9\% \\ 
\hline
Grade I F1-Score & 86.0\% & 84.8\% & +1.2\% \\ 
\hline
Grade II F1-Score   & 80.8\% & 76.1\% & +4.7\% \\ 
\hline
Grade III F1-Score   & 83.2\% & 82.2\% & +1.0\% \\ 
\hline
Best Validation   & 77.6\% & 83.1\% & -5.5\% \\ 
\hline
\end{tabular}
\end{table}

\begin{figure}
\centering
\includegraphics[width=0.8\linewidth]{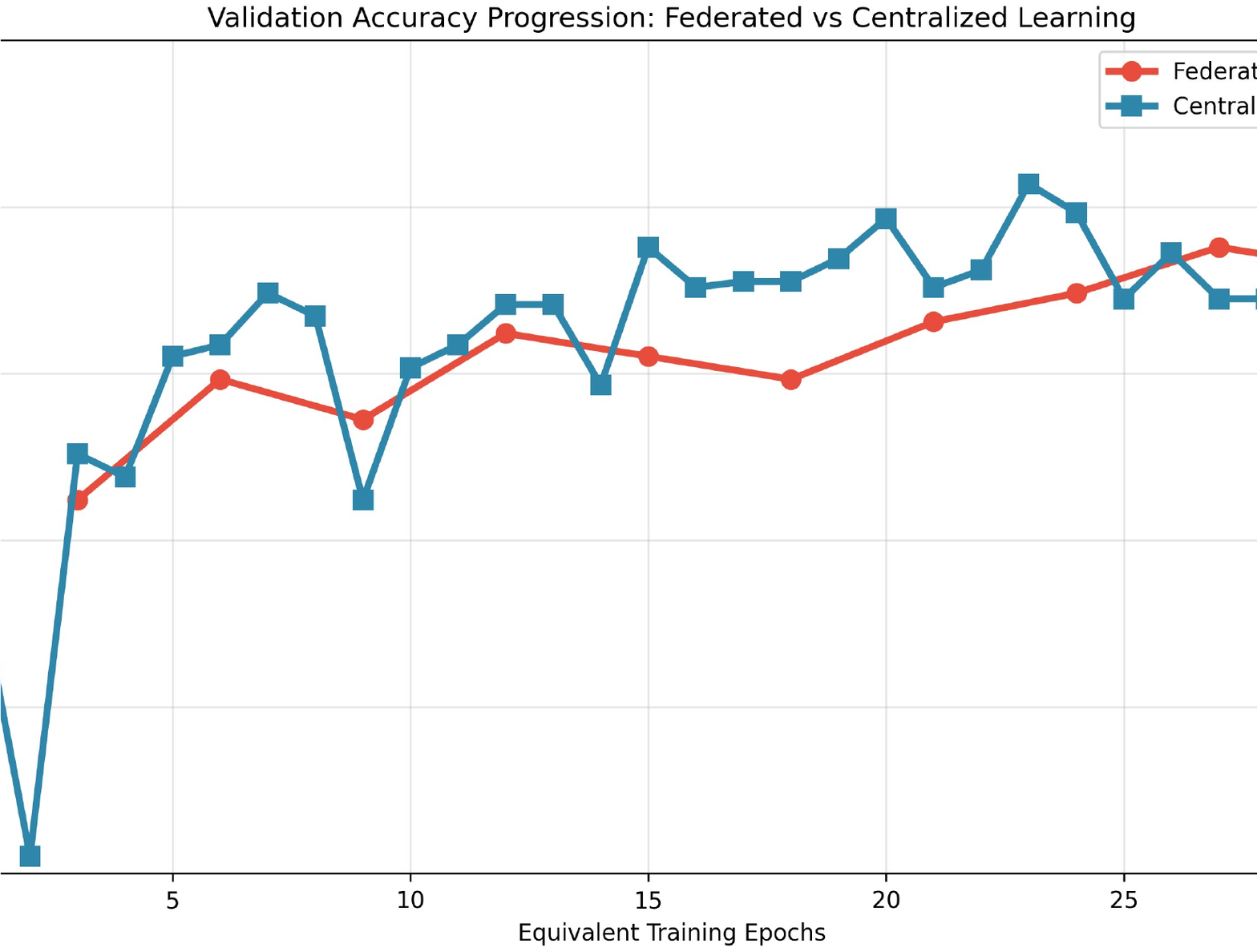}
\caption{Training progression comparison: Federated Learning versus centralized training across validation rounds}
\label{fig:fl_vs_nofl_progression}
\end{figure}

\subsection{Grade-wise Analysis}
The comprehensive description by histopathological grade makes it clear how clinically significant the suggested approach is. The system achieved {92.7\% precision} and {80.2\% recall} for Grade~I (well-differentiated) tumors, with an F1-score of {86.0\%}. This indicates that, even though lacking some genuine Grade~I cases, the approach is highly dependable in identifying cases with less clinical significance and generates a comparatively small number of false positives. Clinically speaking, this compromise is acceptable because Grade~I cancers usually possess a long life expectancy.

The model utilized both sensitivity ({86.0\% recall}) and precision ({76.2\%}) for Grade~II (moderately differentiated) tumors, achieving an F1-score of {80.8\%}. Despite the challenging morphological intersection between well- and poorly differentiated classes, this performance shows that the system can successfully record intermediate cases.  

The case of Grade~III (poorly differentiated) tumors is the most clinically significant. Recall, precision, and F1-scores for the model were {87.5\%}, {79.2\%}, and {83.2\%}, respectively. Given that a false-negative in Grade~III may ultimately result in delayed cancer treatment, the higher recall observed here is the most clinically relevant. In practice, this compromise is preferable since it is less dangerous to be prematurely diagnosed than misdiagnosed, even though the system may occasionally overpredict due to reduced precision. Accordingly, these results highlight the framework’s utility in minimizing clinically dangerous false negatives while still maintaining strong overall accuracy.
These results demonstrate that the system firmly supports the highest clinical grade in terms of sensitivity, patient safety, and diagnostic reliability requirements. This shows the effectiveness of the proposed system in the most clinically relevant grades; therefore, it aligns well with patient safety and diagnostic reliability standards.

\begin{table*}
\centering
\caption{Grade-wise classification performance showing precision, recall, and F1-scores}
\begin{tabular}{ccccc}
\hline
\textbf{Grade} & \textbf{Precision} & \textbf{Recall} & \textbf{F1-Score} & \textbf{Clinical Significance} \\ \hline
I (Well)      & 92.7\% & 80.2\% & 86.0\% & Least critical \\ \hline
II (Moderate) & 76.2\% & 86.0\% & 80.8\% & Intermediate  \\ \hline
III (Poor)    & 79.2\% & 87.5\% & 83.2\% & Most critical  \\ \hline
\end{tabular}
\end{table*}

\begin{figure}[H]
\centering
\includegraphics[width=\linewidth]{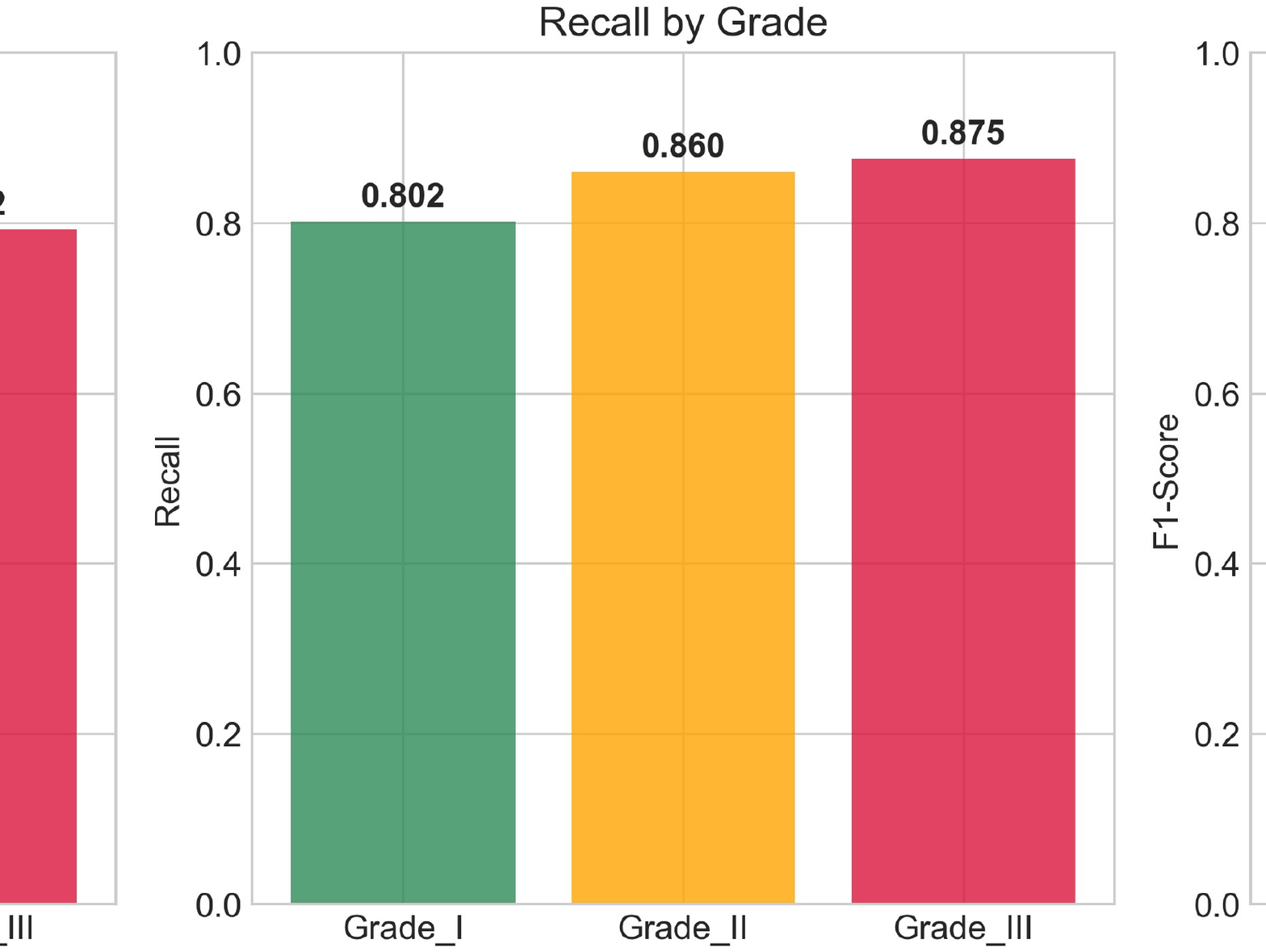}
\caption{Grade-specific performance metrics demonstrating balanced classification across all tumor grades}
\label{fig:grade_wise_analysis}
\end{figure}

\subsection{Magnification Impact}
\begin{table}[H]
\centering
\caption{Model accuracy across different magnification levels in histopathology imaging}
\begin{tabular}{ccc}
\hline
\textbf{Magnification} & \textbf{Accuracy} & \textbf{Clinical Interpretation} \\ \hline
40$\times$ & 88.0\% & Strong nuclear detail (highest accuracy) \\ \hline
20$\times$ & 83.3\% & Tissue architecture \\ \hline
10$\times$ & 83.3\% & Intermediate-level context \\ \hline
4$\times$  & 73.2\% & Global tissue overview (weakest) \\ \hline
\end{tabular}
\end{table}

\begin{figure}[H]
\centering
\includegraphics[width=0.7\linewidth]{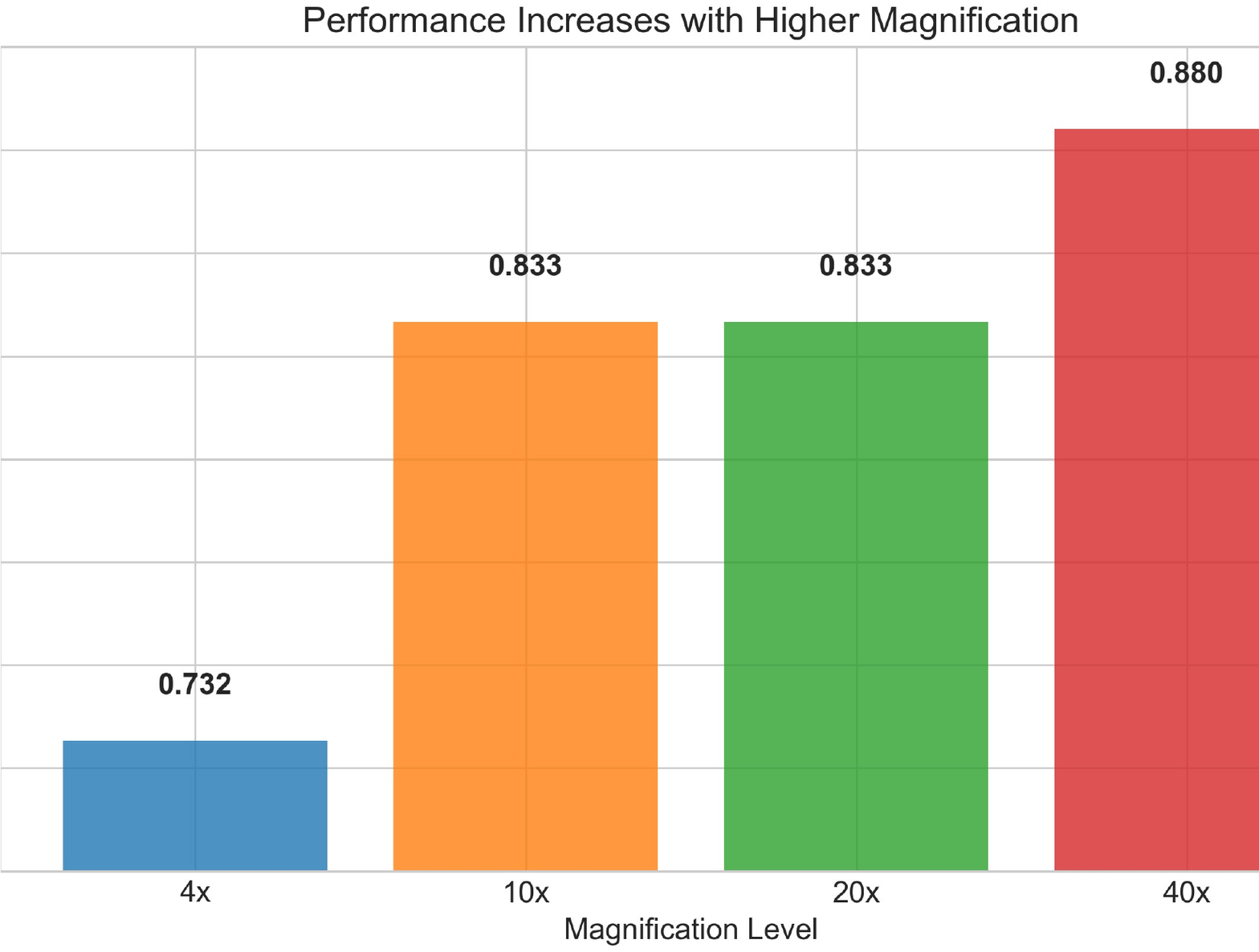}
\caption{Classification accuracy by magnification level, showing performance improvement with increasing resolution}
\label{fig:magnification_acuracy}
\end{figure}

The dual-stream ResNetRS50 backbone's multi-scale stability has been shown by the importance of input magnification in classification performance. Given the rich cell and subcellular details that enable precise grade differentiation, it is expected that the model would capture the greatest amount of information and achieve the highest accuracy of {88.0\%} at 40$\times$ magnification. With values of {83.3\%} at 20$\times$ and 10$\times$ magnifications, respectively, the model also provided good accuracy. This demonstrates that, despite being less informative than nuclear detail, tissue-level architecture and surrounding morphology are still helpful grading features. Interestingly, the 10$\times$ performance was similar to the 20$\times$ result, illustrating the model's ability to leverage lower-resolution intermediate contextual cues.  

The lowest accuracy, however, was achieved at 4$\times$ magnification ({73.2\%}), showing the limitations of global tissue overview images, which lack the fine-grained morphological signals necessary for accurate classification. Clinically, these results highlight the ideal scenario where optimal grading performance can be obtained using both high-resolution (nuclear detail) and intermediate-resolution (tissue context) views. These findings validate a dual-stream foundation, both shallow and deep, that integrates contextual and fine features to ensure scale-invariant performance.  

Combining these results shows that the suggested federated framework may address the essential clinical problem of accurately identifying aggressive Grade~III tumors in addition to achieving high overall performance.
Furthermore, the dependence on magnification illustrates the importance of multi-scale modeling in histopathology, while the federated setting facilitates data transfer and patient privacy protection across hospital networks. The system's potential as a tool for assisting pathologists in grading colorectal cancer in clinical practice is evident in the impressive balance between robustness, generalizability, and clinical safety features.  

\subsection{Training Dynamics}
The model plateaued at Round~9, suggesting an optimal stopping point. FedProx minimized client divergence, ensuring smooth convergence.  

\begin{figure}[H]
\centering
\includegraphics[width=0.8\linewidth]{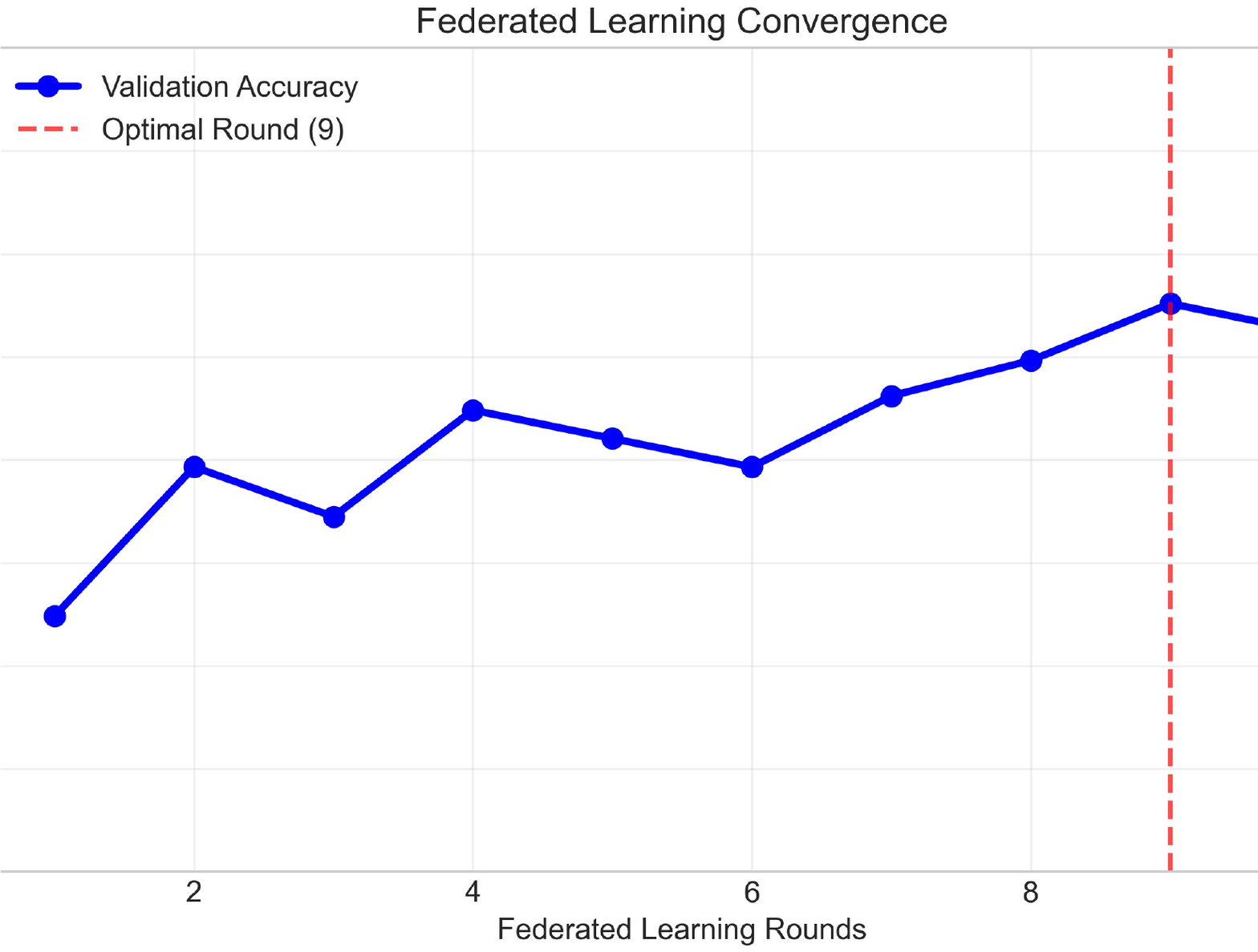}
\caption{Federated Learning convergence profile over 10 training rounds, showing stable improvement to 77.6\% validation accuracy}
\label{fig:training_dynamics}
\end{figure}

\section{Future Work}
Future research should focus on advancing multi-hospital federated learning (FL) networks in real-world clinical settings to validate efficacy across diverse institutions and demographics. Additionally, developing federated domain adaptation techniques to address variations in scanner equipment, staining protocols, and data distribution will improve robustness. To ensure practical utility, FL systems must be integrated into pathology workflows with pathologist-in-the-loop validation. Furthermore, incorporating explainable AI (XAI) features, such as heatmaps and attention maps, into federated frameworks is essential for clinical adoption, as these tools enhance transparency, build trust, and ultimately improve patient outcomes.

\section{Conclusion}


This study presents a federated multi-scale deep learning framework for colorectal cancer grading that integrates coarse tissue and fine cellular features within a privacy-preserving setup. The model achieved 83.5\% accuracy and 87.5\% recall for Grade III tumors, demonstrating strong diagnostic reliability while maintaining data confidentiality. Its modular design with standardized preprocessing, checkpointing, and metadata traceability, that supports scalability across institutions. Future work will expand to larger, more diverse datasets and incorporate molecular data to advance real-world clinical deployment.

\bibliographystyle{plainnat}
\bibliography{Ref-CI-2021}
\end{document}